\documentclass[11pt,letterpaper]{article}
\usepackage{emnlp2016}
\usepackage{times}
\usepackage{url}
\usepackage{latexsym}

\usepackage{amsmath}
\usepackage{amssymb}
\usepackage{booktabs}
\usepackage{graphicx}
\usepackage{multirow}
\usepackage{xcolor}
\usepackage{hyperref}
\emnlpfinalcopy

\title{Improving LSTM-based Video Description \\ with Linguistic Knowledge Mined from Text}

\author{
\hspace{0.8cm}Subhashini Venugopalan \\
\hspace{0.7cm}  UT Austin \\
\hspace{0.6cm} {\tt \small vsub@cs.utexas.edu} \\
  \And
 \hspace{1.2cm} Lisa Anne Hendricks \\
  \hspace{1.2cm}UC Berkeley \\
  \hspace{1.2cm} {\tt \footnotesize lisa\_anne@berkeley.edu} \\
  \And
  Raymond Mooney\hspace{-1.2cm} \\
  UT Austin\hspace{-1.2cm} \\
   \hspace{1.2cm} {\tt \footnotesize mooney@cs.utexas.edu} \\
  \And
  Kate Saenko \hspace{-0.8cm}\\
  Boston University \hspace{-0.7cm}\\
{\tt \footnotesize saenko@bu.edu}\hspace{0.1cm} \\
}

\begin{document}
\maketitle
\begin{abstract}
This paper investigates how linguistic knowledge mined from large text corpora can aid the generation of natural language descriptions of videos. Specifically, we integrate both a neural language model and distributional semantics trained on large text corpora into a recent LSTM-based architecture for video description. We evaluate our approach on a collection of Youtube videos as well as two large movie description datasets showing significant improvements in grammaticality while modestly improving descriptive quality.
\end{abstract}

\section{Introduction}

The ability to automatically describe videos in natural language (NL) enables many important applications including content-based video retrieval and video description for the visually impaired. 
The most effective recent methods \cite{S2VT,yao15arxiv} use recurrent neural networks (RNN) and treat the problem as machine translation (MT) from video to natural language.
Deep learning methods such as RNNs need large training corpora;  however, there is a lack of high-quality paired video-sentence data. In contrast, raw text corpora are widely available and exhibit rich linguistic structure that can aid video description. Most work in statistical MT utilizes both a language model trained on a large corpus of monolingual target language data as well as a translation model trained on more limited parallel bilingual data. 
This paper explores methods to incorporate knowledge from language corpora to capture general linguistic regularities to aid video description.

This paper integrates linguistic information into a video-captioning model based on Long Short Term Memory (LSTM) \cite{schmidLSTM} RNNs which have shown state-of-the-art performance on  the task. Further, LSTMs are also effective as language models (LMs) \cite{sundermeyer:interspeech10}. 
Our first approach (early fusion) 
is to pre-train the network on plain text before training on parallel video-text corpora. 
Our next two approaches, inspired by recent MT work \cite{gulcehreArxiv15}, integrate an LSTM LM with the existing video-to-text model.
Furthermore, we also explore replacing the standard one-hot word encoding with distributional vectors trained on external corpora. 

We present detailed comparisons between the approaches, evaluating them on a standard Youtube corpus and two recent large movie description datasets.
The results demonstrate significant improvements in grammaticality of the descriptions (as determined by crowdsourced human evaluations) and more modest improvements in descriptive quality (as determined by both crowdsourced human judgements and standard automated comparison to human-generated descriptions). Our main contributions are 
1) multiple ways to incorporate knowledge from external text into an existing captioning model, 
2) extensive experiments comparing the methods on three large video-caption datasets, and 
3) human judgements to show that external linguistic knowledge has a significant impact on grammar.
\section{LSTM-based Video Description}
\label{sec:background}
We use the successful S2VT video description framework from \newcite{S2VT} as our underlying model and describe it briefly here.
S2VT uses a sequence to sequence approach \cite{ilya:nips14,cho2014properties} that maps an input $\vec{x} = (x_1, \ldots, x_T)$  video frame feature sequence to  a fixed dimensional vector and then decodes this into a sequence of output words $\vec{y} =(y_1, \ldots, y_N)$.

\begin{figure}[t]
\begin{center}
\includegraphics[width=\linewidth]{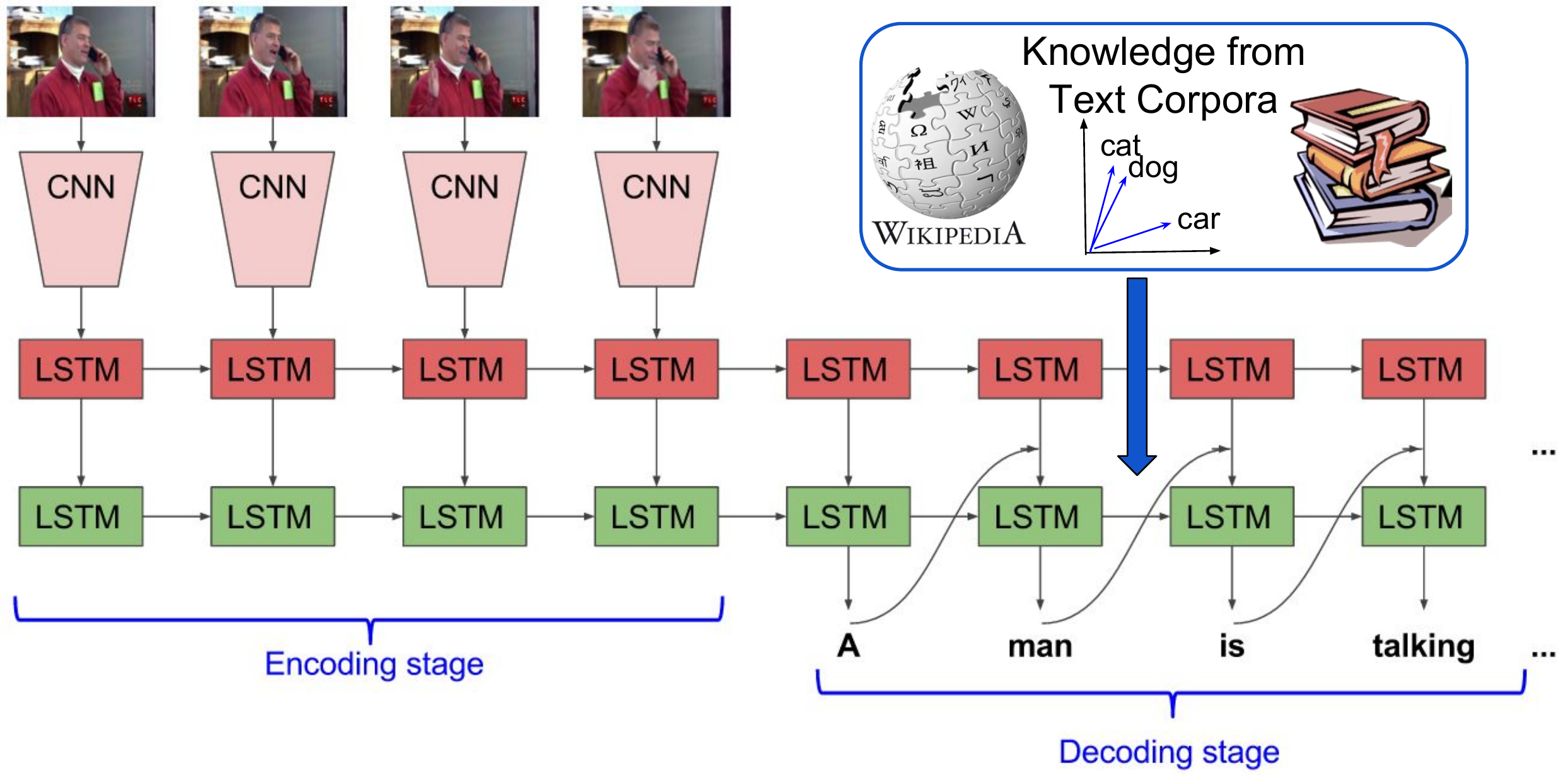}
\end{center} 
 \caption{The S2VT architecture encodes a sequence of frames and decodes them to a sentence. We propose to add knowledge from text corpora to enhance the quality of video description.}
\label{fig:acl-teaser}
\end{figure}

As shown in Fig.\ \ref{fig:acl-teaser}, it employs a stack of two LSTM layers. 
The input $\vec{x}$ to the first LSTM layer is a sequence of frame features obtained from the penultimate layer (fc$_7$) of a Convolutional Neural Network (CNN) after the ReLu operation. This LSTM layer encodes the video sequence. At each time step, the hidden control state $h_t$ is provided as input to a second LSTM layer. After viewing all the frames, the second LSTM layer learns to decode this state into a sequence of words. This can be viewed as using one LSTM layer to model the visual features, and a second LSTM layer to model language conditioned on the visual representation. We modify this architecture to incorporate linguistic knowledge at different stages of the training and generation process. Although our methods use S2VT, they are sufficiently general and could be incorporated into other CNN-RNN based captioning models.

\section{Approach}
\label{sec:approach}

Existing visual captioning models \cite{vinyals14arxiv,donahue15cvpr} are trained solely on text from the caption datasets and tend to exhibit some linguistic irregularities associated with a restricted language model and a small vocabulary. Here, we investigate several techniques to integrate prior linguistic knowledge into a CNN/LSTM-based network for video to text (S2VT) and evaluate their effectiveness at improving the overall description.

\paragraph{Early Fusion.}
Our first approach ({\it early fusion}), is to pre-train portions of the network modeling language on large corpora of raw NL text and then continue ``fine-tuning'' the parameters on the paired video-text corpus.
An LSTM model learns to estimate the probability of an output sequence given an input sequence. To learn a language model, we train the LSTM layer to predict the next word given the previous words. Following the S2VT architecture, we embed one-hot encoded words in lower dimensional vectors. The network is trained on web-scale text corpora and the parameters are learned through backpropagation using stochastic gradient descent.\footnote{The LM was trained to achieve a perplexity of 120}  The weights from this network are then used to {\it initialize} the embedding and weights of the LSTM layers of S2VT, which is then trained on video-text data. This trained LM is also used as the LSTM LM in the late and deep fusion models.

\paragraph{Late Fusion.}
Our late fusion approach is similar to how neural machine translation models incorporate a trained language model during decoding. At each step of sentence generation, the video caption model proposes a distribution over the vocabulary. We then use the language model to re-score the final output by considering the weighted average of the sum of scores proposed by the LM as well as the S2VT video-description model (VM). More specifically, if $y_t$ denotes the output at time step $t$, and if $p_{VM}$ and $p_{LM}$ denote the proposal distributions of the video captioning model, and the language models respectively, then for all words $y' \in V$ in the vocabulary  we can recompute the score of each new word, $p(y_t = y')$ as:
\vspace{-2.0mm}
\begin{equation}
\alpha \cdot p_{VM}(y_t = y') + (1 - \alpha) \cdot p_{LM} (y_t = y')
\vspace{-2.0mm}
\end{equation}
Hyper-parameter $\alpha$ is tuned on the validation set.

\begin{figure}[t]
\includegraphics[width=1\linewidth]{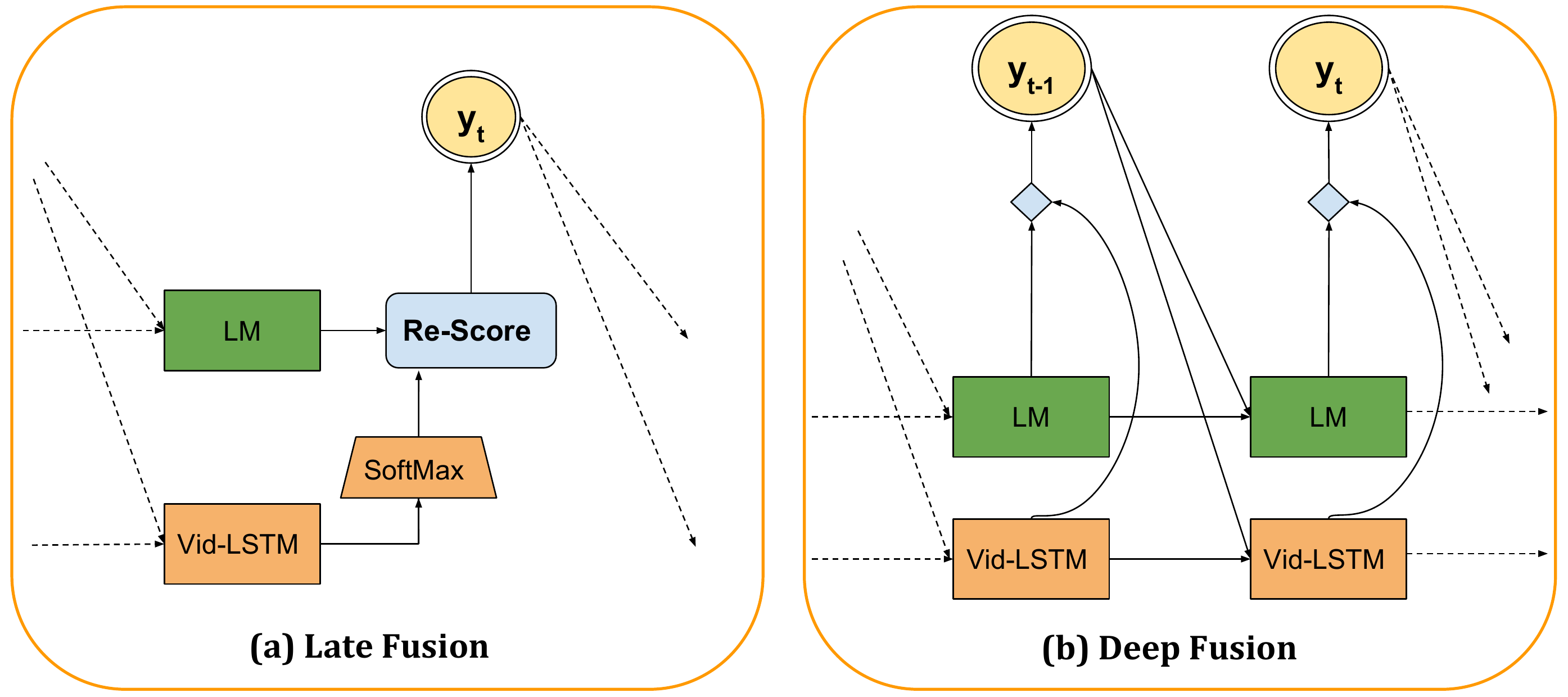}
\caption{
 \small
Illustration of our late and deep fusion approaches to integrate an independently trained LM to aid video captioning. The deep fusion model learns jointly from the hidden representations of the LM and S2VT video-to-text model (Vid-LSTM), whereas the late fusion re-scores the softmax output of the video-to-text model.
}
\label{fig:fusion}
\end{figure}

\paragraph{Deep Fusion.}
In the deep fusion approach (Fig.~\ref{fig:fusion}), we integrate the LM a step deeper in the generation process by concatenating the hidden state of the language model LSTM ($h_t^{LM}$) with the hidden state of the S2VT video description model ($h_t^{VM}$) and use the combined latent vector to predict the output word. This is similar to the technique proposed by \newcite{gulcehreArxiv15} for incorporating language models trained on monolingual corpora for machine translation. However, our approach differs in two key ways: (1) we only concatenate the hidden states of the S2VT LSTM and language LSTM and do not use any additional context information, (2)  we fix the weights of the LSTM language model but train the full video captioning network. 
In this case, the probability of the predicted word at time step $t$ is:
\vspace{-3.0mm}
\begin{equation}
p(y_t | \vec{y}_{<t}, \vec{x}) \propto  \mathrm{exp}(\mathrm{W f}(h_t^{VM}, h_t^{LM}) + b)
\vspace{-2.0mm}
\end{equation}
where $\vec{x}$ is the visual feature input, $W$ is the weight matrix, and $b$ the biases.
We avoid tuning the LSTM LM to prevent overwriting already learned weights of a strong language model. But we train the full video caption model to incorporate the LM outputs while training on the caption domain.

\paragraph{Distributional Word Representations.}
\label{sec:glove_init}
The S2VT network, like most image and video captioning models, represents words using a 1-of-N (one hot) encoding. During training, the model learns to embed ``one-hot'' words into a lower 500d space by applying a linear transformation. However, the embedding is learned only from the limited and possibly noisy text in the caption data. There are many approaches \cite{Mikolov,Glove} that use large text corpora to learn vector-space representations of words that capture fine-grained semantic and syntactic regularities. We propose to take advantage of these to aid video description.  Specifically, we replace the embedding matrix from one-hot vectors and instead use 300-dimensional GloVe vectors \cite{Glove} pre-trained on 6B tokens from Gigaword and Wikipedia 2014. In addition to using the distributional vectors for the input, we also explore variations where the model predicts both the one-hot word (trained on the $\mathrm{softmax}$ loss), as well as predicting the distributional vector from the LSTM hidden state using  Euclidean loss as the objective. Here the output vector ($y_t$) is computed as $y_t = (W_g h_{t} + b_g)$, and the loss is given by:
\vspace{-1.0mm}
\begin{align}
\mathbb{L}(y_t, w_{glove}) &= \| (W_g h_{t} + b_g) - w_{glove} \| ^{2}
\vspace{-1.0mm}
\end{align}
where $h_{t}$ is the LSTM output, $w_{glove}$ is the word's GloVe embedding and $W$, $b$ are weights and biases. The network then essentially becomes a multi-task model with two loss functions. However, we use this loss only to influence the weights learned by the network, the predicted word embedding is not used.

\paragraph{Ensembling.}
The overall loss function of the video-caption network is non-convex, and difficult to optimize. In practice, using an ensemble of networks trained slightly differently can improve performance  \cite{hansen90ensemble}. In our work we also present results of an ensemble by averaging the predictions of the best performing models.

\section{Experiments}

\paragraph{Datasets.} 
Our language model was trained on sentences from Gigaword, BNC, UkWaC, and Wikipedia. The vocabulary consisted of 72,700 most frequent tokens also containing GloVe embeddings. Following the evaluation in \newcite{S2VT}, we compare our models on the Youtube dataset~\cite{chen11acl}, as well as two large movie description corpora: MPII-MD~\cite{rohrbach15cvpr} and M-VAD~\cite{torabi15arxiv}.

\paragraph{Evaluation Metrics.} We evaluate performance using machine translation (MT) metrics METEOR~\cite{banerjee2005meteor} and BLEU \cite{papineni2002bleu} to compare the machine-generated descriptions to human ones. For the movie corpora which have just a single description we use only METEOR which is more robust.

\paragraph{Human Evaluation.} We also obtain human judgements using Amazon Turk on a random subset of 200 video clips for each dataset. Each sentence was rated by 3 workers on a Likert scale of 1 to 5 (higher is better) for relevance and grammar. No video was provided during grammar evaluation. For movies, due to copyright, we only evaluate on grammar. %

\vspace{-0.2cm}
\subsection{Youtube Video Dataset Results}
\vspace{-0.2cm}
Comparison of the proposed techniques in Table~\ref{tab:yt} shows that Deep Fusion performs well on both METEOR and BLEU; incorporating Glove embeddings substantially increases METEOR, and combining them both does best. Our final model is an ensemble (weighted average) of the Glove, and the two Glove+Deep Fusion models trained on the external and in-domain COCO~\cite{coco2014} sentences. We note here that the state-of-the-art on this dataset is achieved by HRNE~\cite{pan15arxiv} (METEOR 33.1) %
which proposes a superior visual processing pipeline using attention to encode the video.

Human ratings also correlate well with the METEOR scores, confirming that our methods give a modest improvement in descriptive quality.  However, incorporating linguistic knowledge significantly\footnote{Using the Wilcoxon Signed-Rank test, results were significant with $p<0.02$ on relevance and $p<0.001$ on grammar.} improves the grammaticality of the results, making them more comprehensible to human users.

\vspace{-0.3cm}
\paragraph{Embedding Influence.} We experimented multiple ways to incorporate word embeddings: {\it (1) GloVe input:} Replacing one-hot vectors with GloVe on the LSTM input performed best.
{\it (2) Fine-tuning:} Initializing with GloVe and subsequently fine-tuning the embedding matrix reduced  validation results by 0.4 METEOR. {\it  (3) Input and Predict.} Training the LSTM to accept and predict GloVe vectors, as described in Section \ref{sec:glove_init}, performed similar to (1). All scores reported in Tables \ref{tab:yt} and \ref{tab:results:movies} correspond to the setting in (1) with GloVe embeddings only as input.

\begin{table}[t]
\small
\begin{center}
\setlength\tabcolsep{4.0pt}
\begin{tabular}{lcccl}
\toprule
Model & {\footnotesize{METEOR}} & {\footnotesize{B-4}} & {\footnotesize{Relevance}} & {\footnotesize{Grammar}}\\
\cmidrule(lr){1-1}\cmidrule(lr){2-2}\cmidrule(lr){3-3}\cmidrule(lr){4-4}\cmidrule(lr){5-5}
S2VT & 29.2 & 37.0 & 2.06 & 3.76 \\
Early Fusion & {29.6} & 37.6 & - & -\\
Late Fusion & 29.4 & 37.2 & - & -\\
Deep Fusion & {29.6} & {39.3} & - & -\\
Glove & {30.0} & {37.0} & - & -\\
\cmidrule(lr){1-1}\cmidrule(lr){2-2}\cmidrule(lr){3-3}\cmidrule(lr){4-4}\cmidrule(lr){5-5}
Glove+Deep \\
\ - Web Corpus & {30.3} & 38.1 & 2.12 \ \quad & 4.05*\\
\ - In-Domain & 30.3 & 38.8 & 2.21* & 4.17*\\
Ensemble & {\bf31.4} & {\bf42.1} & \textbf{2.24}* & \textbf{4.20}* \\
\cmidrule(lr){1-1}\cmidrule(lr){2-2}\cmidrule(lr){3-3}\cmidrule(lr){4-4}\cmidrule(lr){5-5}
Human & - & - & 4.52 & 4.47 \\
\bottomrule
\end{tabular}
\end{center}
\caption{\small Youtube dataset: METEOR and BLEU@4 in $\%$, and human ratings (1-5) on relevance and grammar. Best results in bold, * indicates significant over S2VT.}
\label{tab:yt}
\end{table}

\begin{table}
\small
\begin{center}
\setlength\tabcolsep{4.0pt}
\begin{tabular}{l|cc|cc}
\toprule
Model & \multicolumn{2}{|c|}{MPII-MD} & \multicolumn{2}{|c}{M-VAD} \\ \cline{2-5}
 & {\footnotesize{METEOR}} & {\footnotesize{Grammar}}  & {\footnotesize{METEOR}} & {\footnotesize{Grammar}} \\ 
\hline
\rule{0pt}{4ex} S2VT$^{\dagger}$ & 6.5 & 2.6 & 6.6 & 2.2 \\
Early Fusion & 6.7 & - & 6.8 & -\\
Late Fusion & 6.5 & - & 6.7 & -\\
Deep Fusion & 6.8 & - & 6.8 & -\\
Glove & 6.7 & {\ \ 3.9}* & 6.7 & {\ \ 3.1}*\\
Glove+Deep & 6.8 & {\ \ \bf{4.1}}* & 6.7 & {\ \ \bf{3.3}}* \\
\bottomrule
\end{tabular}
\end{center}
\caption{\small
Movie Corpora: METEOR (\%) and human grammar ratings (1-5, higher is better). Best results in bold, * indicates significant over S2VT.
}
\label{tab:results:movies}
\vspace{-2.0mm}
\end{table}

\subsection{Movie Description Results}
\begin{figure}
\begin{center}
\includegraphics[width=\linewidth]{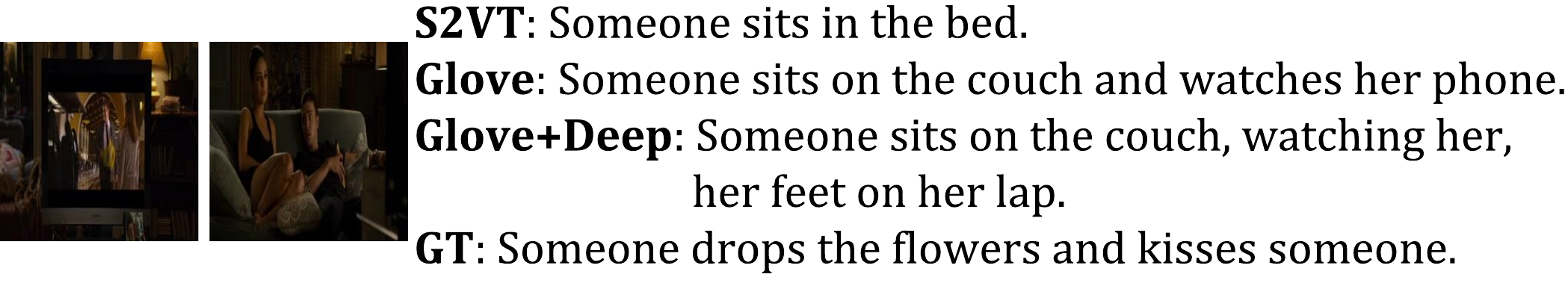}
\end{center}
\vspace{-0.3cm}
 \caption{\small 
Two frames from a clip. Models generate visually relevant sentences but differ from groundtruth (\footnotesize{GT}).}
\label{fig:movie_qual}
\end{figure}

Results on the movie corpora are presented in Table \ref{tab:results:movies}. Both MPII-MD and M-VAD have only a single ground truth description for each video, which makes both learning and evaluation very challenging (E.g. Fig.\ref{fig:movie_qual}). METEOR scores are fairly low on both datasets since generated sentences are compared to a single reference translation.
S2VT$^{\dagger}$ is a re-implementation of the base S2VT model with the new vocabulary and architecture (embedding dimension).
We observe that the ability of external linguistic knowledge to improve METEOR scores on these challenging datasets is small but consistent. Again, human evaluations show significant (with $p < 0.0001$) improvement in grammatical quality.

\vspace{-0.3cm}
\section{Related Work}
\label{sec:related}
\vspace{-0.1cm}
Following the success of LSTM-based models on Machine Translation \cite{ilya:nips14,bahdanau2014neural}, and image captioning \cite{vinyals14arxiv,donahue15cvpr}, recent 
 video description works \cite{venugopalan15naacl,S2VT,yao15arxiv} propose CNN-RNN based models that generate a vector representation for the video and ``decode'' it using an LSTM sequence model to generate a description.
\newcite{venugopalan15naacl} also incorporate external data such as images with captions to improve video description, however in this work, our focus is on integrating external linguistic knowledge for video captioning. We specifically investigate the use of distributional semantic embeddings and LSTM-based language models trained on external text corpora to aid existing CNN-RNN based video description models.

LSTMs have proven to be very effective language models \cite{sundermeyer:interspeech10}. \newcite{gulcehreArxiv15} developed an LSTM model for machine translation that incorporates a monolingual language model for the target language showing improved results. We utilize similar approaches (late fusion, deep fusion) to train an LSTM for translating video to text that exploits large monolingual-English corpora (Wikipedia, BNC, UkWac) to improve RNN based video description networks. However, unlike \newcite{gulcehreArxiv15} where the monolingual LM is used only to tune specific parameters of the translation network, the key advantage of our approach is that the output of the monolingual language model is used (as an input) when training the full underlying video description network.

Contemporaneous to us, \newcite{yu15arxiv}, \newcite{pan15arxiv} and \newcite{ballas15arxiv} propose video description models focusing primarily on improving the video representation itself using a hierarchical visual pipeline, and attention. Without the attention mechanism their models achieve METEOR scores of 31.1, 32.1 and 31.6 respectively on the Youtube dataset. The interesting aspect, as demonstrated in our experiments (Table \ref{tab:yt}), is that the contribution of language alone is considerable and only slightly less than the visual contribution on this dataset. Hence, it is important to focus on both aspects to generate better descriptions. %

\vspace{-0.3cm}
\section{Conclusion}
This paper investigates multiple techniques to incorporate linguistic knowledge from text corpora to aid video captioning. We empirically evaluate our approaches on Youtube clips as well as two movie description corpora. Our results show significant improvements on human evaluations of grammar while modestly improving the overall descriptive quality of sentences on all datasets. 
While the proposed techniques are evaluated on a specific video-caption network,
they are generic and can be applied to other video and image captioning
models \cite{hendricks16cvpr,venugopalan16noc}. The code and models are shared on { \href{http://vsubhashini.github.io/language\_fusion.html}{\url{http://vsubhashini.github.io/language\_fusion.html}}}.

\begin{figure}[t!]
\begin{center}
\includegraphics[width=\linewidth]{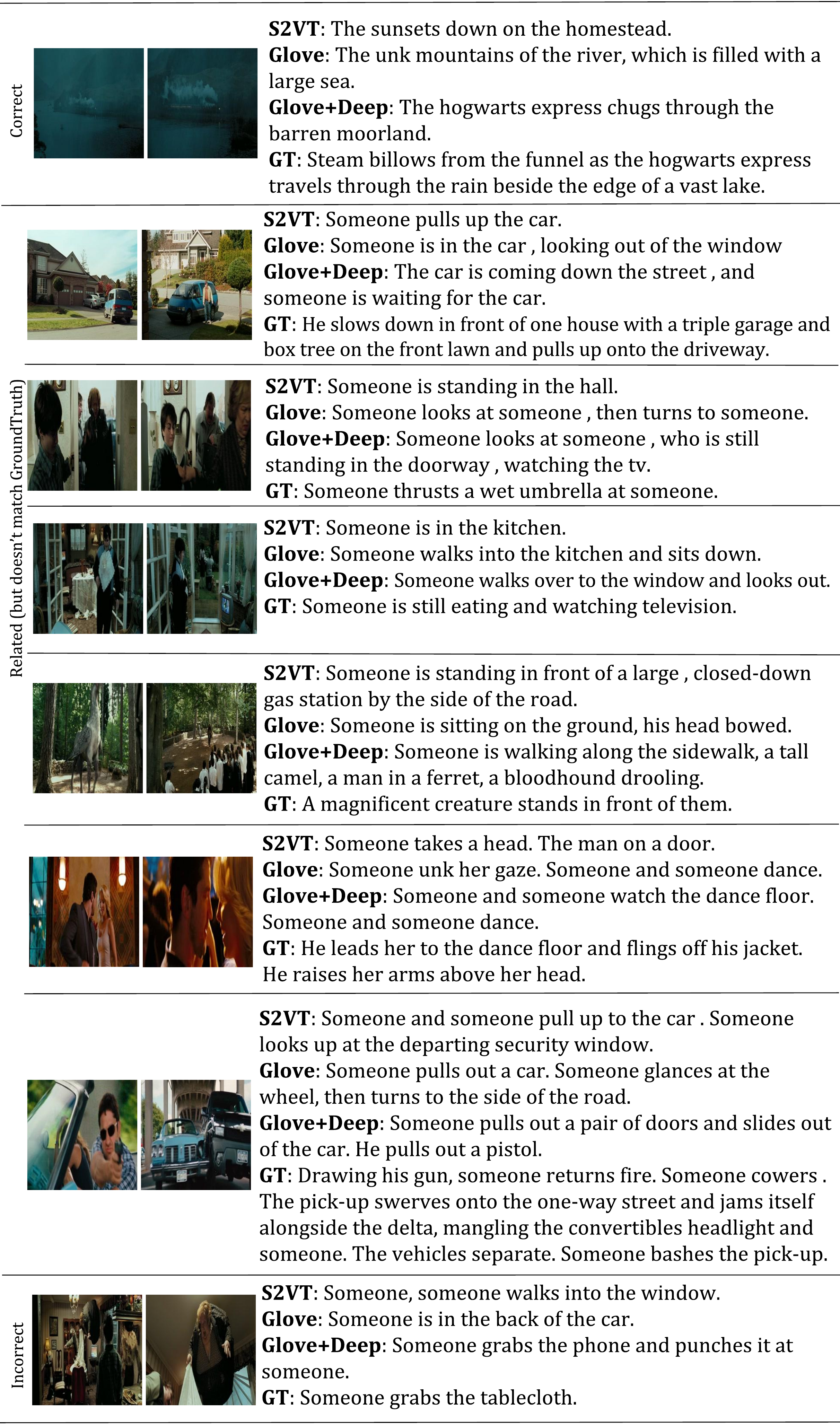}
\end{center}
\vspace{-0.2cm}
 \caption{Representative frames from clips in the movie description corpora. S2VT is the baseline model, Glove indicates the model trained with input Glove vectors, and Glove+Deep uses input Glove vectors with the Deep Fusion approach. GT indicates groundtruth sentence.}
\label{fig:movie_more_ex}
\end{figure}

\section*{Acknowledgements}
This work was supported by NSF awards IIS-1427425 and IIS-1212798, and ONR ATL Grant N00014-11-1-010, and DARPA under AFRL grant FA8750-13-2-0026. Raymond Mooney and Kate Saenko also acknowledge support from a Google grant. Lisa Anne Hendricks is supported by the National Defense Science and Engineering Graduate (NDSEG) Fellowship.
\clearpage
\bibliographystyle{emnlp2016}
\bibliography{biblioShort,rohrbach,related,refs,acl16refs}

\end{document}